\documentclass[10pt,twocolumn,letterpaper]{article}

\usepackage{cvpr}              

\usepackage[dvipsnames]{xcolor}

\definecolor{cvprblue}{rgb}{0.21,0.49,0.74}
\usepackage[pagebackref,breaklinks,colorlinks,citecolor=cvprblue]{hyperref}
\usepackage{multirow}
\usepackage{wrapfig}
\usepackage{colortbl}
\usepackage{bbding}
\usepackage{comment}
\usepackage{float}
\def\paperID{6124} 
\def\confName{CVPR}
\def\confYear{2024}

\title{BT-Adapter: Video Conversation is Feasible Without Video Instruction Tuning}

\author{{Ruyang Liu}~$^{*1,2 \Diamond}$
    ~~~{Chen Li}~$^{*3}$
    ~~~{Yixiao Ge}~$^{3}$
    ~~~Thomas H. Li$^{1}$ 
    ~~~{Ying Shan}~$^{3}$
    ~~~{Ge Li \footnotesize{\Envelope}}$^{1}$\\
    {\small $^1$School of Electronic and Computer Engineering, Shenzhen Graduate School, Peking University} \\ 
    {\small ~~~~$^2$Peng Cheng Laboratory~~~~$^3$Applied Research Center (ARC), Tencent PCG}\\ 
    {\tt\small \{ruyang@stu,geli@ece,thomas@\}.pku.edu.cn}~~~~{\tt\small\{palchenli,yixiaoge,yingsshan\}@tencent.com}\\  
}

\begin{document}
\maketitle
\begin{abstract}
The recent progress in Large Language Models (LLM) has spurred various advancements in image-language conversation agents, 
while how to build a proficient video-based dialogue system is still under exploration. 
Considering the extensive scale of LLM and visual backbone, minimal GPU memory is left for facilitating effective temporal modeling, which is crucial for comprehending and providing feedback on videos. 
To this end, we propose Branching Temporal Adapter (BT-Adapter), a novel method for extending image-language pretrained models into the video domain. Specifically, BT-Adapter serves as a plug-and-use temporal modeling branch alongside the pretrained visual encoder, which is tuned while keeping the backbone frozen. 
Just pretrained once, BT-Adapter can be seamlessly integrated into all image conversation models using this version of CLIP, enabling video conversations without the need for video instructions.
Besides, we develop a unique asymmetric token masking strategy inside the branch with tailor-made training tasks for BT-Adapter, facilitating faster convergence and better results.
Thanks to BT-Adapter, we are able to empower existing multimodal dialogue models with strong video understanding capabilities without incurring excessive GPU costs.
Without bells and whistles, BT-Adapter achieves (1) state-of-the-art zero-shot results on various video tasks using thousands of fewer GPU hours. (2) better performance than current video chatbots without any video instruction tuning. (3) state-of-the-art results of video chatting using video instruction tuning, outperforming previous SOTAs by a large margin. The code has been available at 
\href{https://github.com/farewellthree/BT-Adapter}{https://github.com/farewellthree/BT-Adapter}.
\end{abstract}    
\section{Introduction}
\label{intro}

\renewcommand{\thefootnote}{ } 
\footnotetext{\Envelope Corresponding Author~~~~~~~* equal contribution}

Since the past year, Large Language Model (LLM) chatbots \cite{touvron2023llama, wang2022self, ouyang2022training, zeng2022glm} have emerged as one of the most remarkable advancements within the AI community. Excitingly, the LLM-driven AI assistants have showcased impressive abilities in comprehension, reasoning, and engaging in conversations. 
The success of text-only dialogue systems has also sparked the development of image-language conversation agents \cite{liu2023visual, dai2023instructblip, zhu2023minigpt, gao2023llama}. These agents combine pretrained image models with LLMs, followed by instruction tuning, to create a fusion of visual and textual understanding for enhanced conversation capabilities.

In comparison to images, videos offer a more comprehensive representation of how humans perceive and interpret the world. Nevertheless, constructing a video-centric dialogue system is a more complex endeavor compared to the image-based one. Firstly, a typically multimodal agent, comprising LLM, a pretrained image encoder (usually CLIP \cite{radford2021learning}), and additional learnable modules, is already demanding in terms of GPU memory. The incorporation of video dialogue introduces even higher costs, as multiple frames need to be fed as inputs. Secondly, a pretrained encoder as potent and knowledge-enriched as CLIP is currently lacking in the realm of videos, leading to inferior visual understanding. Lastly, gathering video-text instruction data of comparable scale, quality, and diversity to images poses a significant challenge. Rather than building the video agent from scratch, an alternative and promising strategy involves adapting existing pretrained image-centric dialogue models to the video domain \cite{li2023videochat, maaz2023video, zhang2023video, luo2023valley}, which can leverage the abundant knowledge embedded within image models.

\begin{figure*}[t] 
    \centering
    \includegraphics[width=1\textwidth]{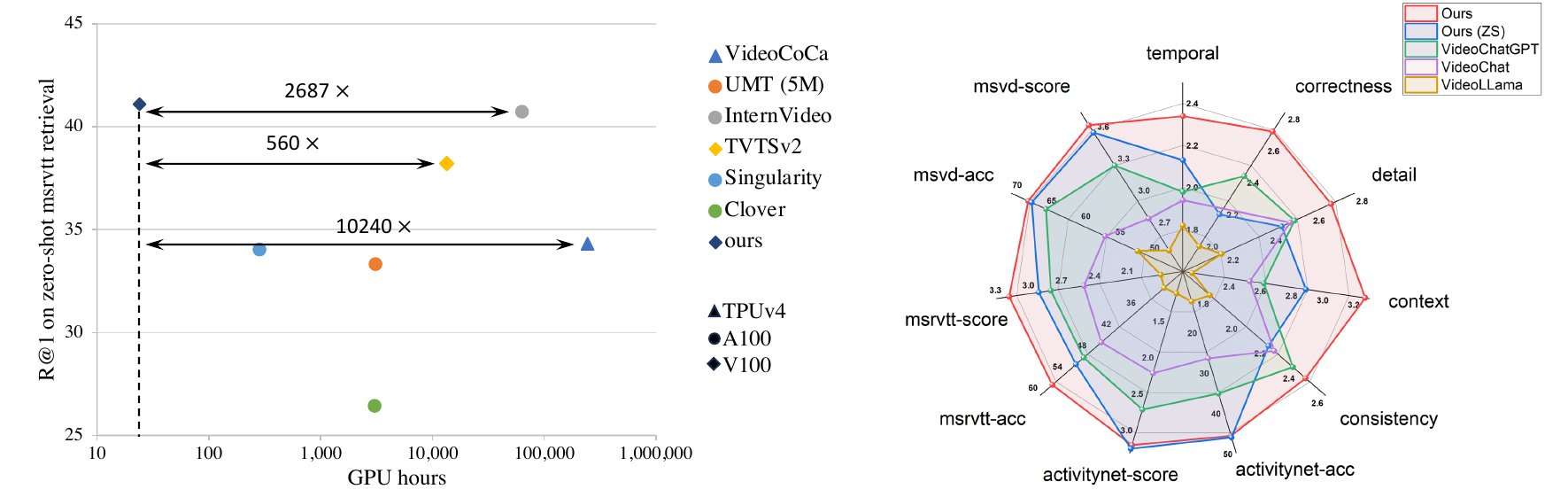} 
    \caption{The performance overview of our BT-Adapter. On the left, we report zero-shot Recall@1 on MSRVTT \cite{xu2016msr} vs. pretraining GPU hours. On the right, we provide a quantitative comparison of video conversations among existing video dialogue agents.}
    \label{image_intro}
    \vspace{-0.5em}
  \end{figure*}

\renewcommand{\thefootnote}{ } 

Expanding pretrained image models to the video domain is an extensively explored area. The crux of the matter lies in enhancing the capability of 2D models to model temporal dynamics \cite{bertasius2021space, carreira2017quo, tran2015learning}. 
However, implementing effective temporal modeling within image dialogue models presents notable challenges, primarily due to the trade-off between efficiency and effectiveness in CLIP-based temporal modeling techniques. Methods with strong temporal modeling capabilities, \textit{e.g.,} joint-ST modeling \cite{bertasius2021space, wang2022internvideo, li2023unmasked, xue2022clip} and interpolate-style modeling \cite{ni2022expanding, pan2022st}, often necessitate finetuning of the entire visual encoder, thereby exacerbating the already significant GPU memory consumption of video conversation models. Moreover, full-finetuning would lead to the loss of knowledge encapsulated in pretrained models, resulting in reduced performance for image-based conversations
In contrast, methods that focus on parameter-efficient temporal modeling aim to keep the image encoder frozen and introduce only a few trainable parameters, which are embraced by most video dialogue models. However, these approaches are typically limited temporal modeling and struggle to capture crucial spatiotemporal features.

To tackle the problems in the aforementioned methods, we proposed Branching Temporal Adapter (BT-Adapter), a novel framework to migrate image-text pretrained models to the video domain. As the name implies, BT-Adapter incorporates a branching spatial-temporal module for temporal modeling. Alongside the pretrained image model, BT-Adapter inherits the parameter efficiency advantage from conventional adapters \cite{houlsby2019parameter} while concurrently achieving effective temporal modeling. Notably, unlike the plug-style adapter, BT-Adapter does not disrupt the forward progression of the pretrained models, thereby safeguarding the integrity of the pretrained multimodal knowledge. After being pretrained with any version of CLIP, BT-Adapter can be seamlessly integrated with all image conversation models using this version of CLIP to activate video conversation, without necessitating video instructions, \textit{e.g.}, openai-CLIP for LLaVa, Eva-CLIP for MiniGPT4 and InstructionBLIP. Additionally, we have devised a unique asymmetric masking mechanism that exclusively implements tube token masking within the BT-Adapter. Building upon this, we formulate two custom training objectives for the BT-Adapter: Masked Branching Token Alignment (MBTA) and Masked Branching Cross-modal Alignment (MBCA). This approach not only reduces computational demands and accelerates convergence but also yields improved outcomes.

To validate the effectiveness and efficiency of our BT-Adapter, we provide a detailed analysis of various temporal modeling strategies for video conversation in Sec. \ref{method_temp}. Furthermore, in Sec. \ref{exp}, we carry out extensive qualitative and quantitative experiments on BT-Adapter, encompassing both traditional video tasks and video conversations.
As depicted in Fig. \ref{image_intro}(left), a series of designs ensures that BT-Adapter is highly resource-efficient: our post-pretraining demands just \textbf{8 V100(32G)} GPUs in a mere \textbf{3 hours}, leading to a reduction in carbon emissions by more than \textbf{10240$\times$} and \textbf{2687$\times$} compared to CoCa \cite{yu2022coca} and InternVideo \cite{wang2022internvideo} respectively. Building upon this, we still achieve state-of-the-art results in zero-shot video-text retrieval. Regarding video conversation, as shown in Fig. \ref{image_intro}(right), we clearly demonstrate that fine-tuned BT-Adapter surpasses the previous state-of-the-art by a significant margin across all benchmarks, and BT-Adapter without instruction tuning has better average performance than fine-tuned SOTAs. 


Finally, Our contributions can be summarized as (1) Comprehensive studies on potential temporal modeling methods for video dialogue systems. (2) Branching Temporal Adapter (BT-Adapter) with tail-made masking and training strategies. (3) State-of-the-art results in conventional video tasks and video conversations.
\section{Related Work}
\vspace{-0.5em}
\noindent \textbf{Temporal Modeling on Image-Language Pretrained Models.}
With the wide application of pretrained image-language models \citep{radford2021learning,yu2022coca}, how to extend these pretrained multimodal models into the video domain has emerged as a novel yet critical research topic. 
Existing temporal modeling methods can be broadly categorized into several types. 
The most straightforward and commonly used one is joint-spatial-temporal modeling \citep{bertasius2021space, wang2022internvideo, li2023unmasked, xue2022clip}. By inputting all video tokens into the image encoder, this approach can effectively model temporal dependencies without other techniques. However, joint-ST modeling necessitates fine-tuning the entire encoder, resulting in significant computational costs and the degradation of pretrained knowledge. 
Another representative type is the interpolate-style temporal modeling, including separated-ST modeling \citep{bertasius2021space, zeng2023tvtsv2}, message token \citep{ni2022expanding}, and ST-Adapter \citep{pan2022st}. Inserting temporal modules between the pretrained spatial layers, interpolate-style modeling shares similar cons and pros with joint-ST modeling. Specifically, ST-Adapter is also known as the parameter-efficient temporal module, where the distinctions with our methods are detailed in Sec. \ref{similar_sec} in the Appendix.
Different from the former two types, concatenate-style temporal modeling \citep{luo2022clip4clip, fang2021clip2video} attaches temporal modules following the pretrained encoder, which allows for the freezing of the backbone and the preservation of knowledge. Nevertheless, this style offers limited temporal modeling, as it hardly captures crucial low-level spatial-temporal features. In Sec. \ref{method_temp}, we will experimentally compare the different temporal modeling methods for video dialogue.

In contrast to the preceding types, BT-Adapter adopts a branching temporal modeling structure. Thanks to the design, we circumvent the issues above, enabling proficient temporal modeling, efficient parameter fine-tuning, and multimodal affinity simultaneously.
Most similar to our methods, STAN \citep{liu2023revisiting} also introduced branch-style temporal modeling. Nonetheless, there are three key distinctions between our work and STAN: (1) Our emphasis is on zero-shot image-to-video transfer and video conversations, areas that have never been explored by STAN. (2) Our performance is notably superior when the backbone is frozen, which is attributed to our unique design like temporal CLS token, zero-initialized temporal projection, and temporal selection. (3) We have introduced the novel training strategy and objectives tailored for the branching temporal structure. 

\noindent \textbf{Video Conversation.} 
Recent advancements in multimodal learning have been predominantly propelled by the fusion of visual models with LLM. \citet{yuan2021florence} initially demonstrated the potential of combining visual models with LLM. Blip-2 \citep{li2023blip} further proposed Q-former that maps visual tokens into the text embedding space. Subsequently, several methods presented visual instruction tuning \citep{liu2023visual,zhu2023minigpt,dai2023instructblip} for image-LLM to enable visual conversation. Most closely related to our topic, there have been several developments in the field of video-centric dialogue models over the past few months \citep{li2023videochat,zhang2023video,maaz2023video,luo2023valley}. 
Generally, these models consist of the visual encoder, LLM, and temporal module, tuned with video instruction data to realize video conversation. However, in pursuit of efficient training, the temporal modules utilized by these models, such as temporal position embedding \citep{zhang2023video} and temporal pooling \citep{maaz2023video,luo2023valley}, often exhibit limited temporal modeling capabilities. In contrast, our approach maintains parameter-efficient training while simultaneously delivering effective temporal modeling. Thanks to the benchmark raised by \citet{maaz2023video}, we can quantitatively illustrate the superiority of our methods in Sec. \ref{exp}. 
Moreover, we introduce a novel setting of zero-shot conversation by integrating pretrained temporal modules with image dialogue models, demonstrating the feasibility of video conversation without any video instruction tuning. 
\section{Methodology} \label{method}
\begin{figure*}[t] 
    \centering
    \includegraphics[width=1\textwidth]{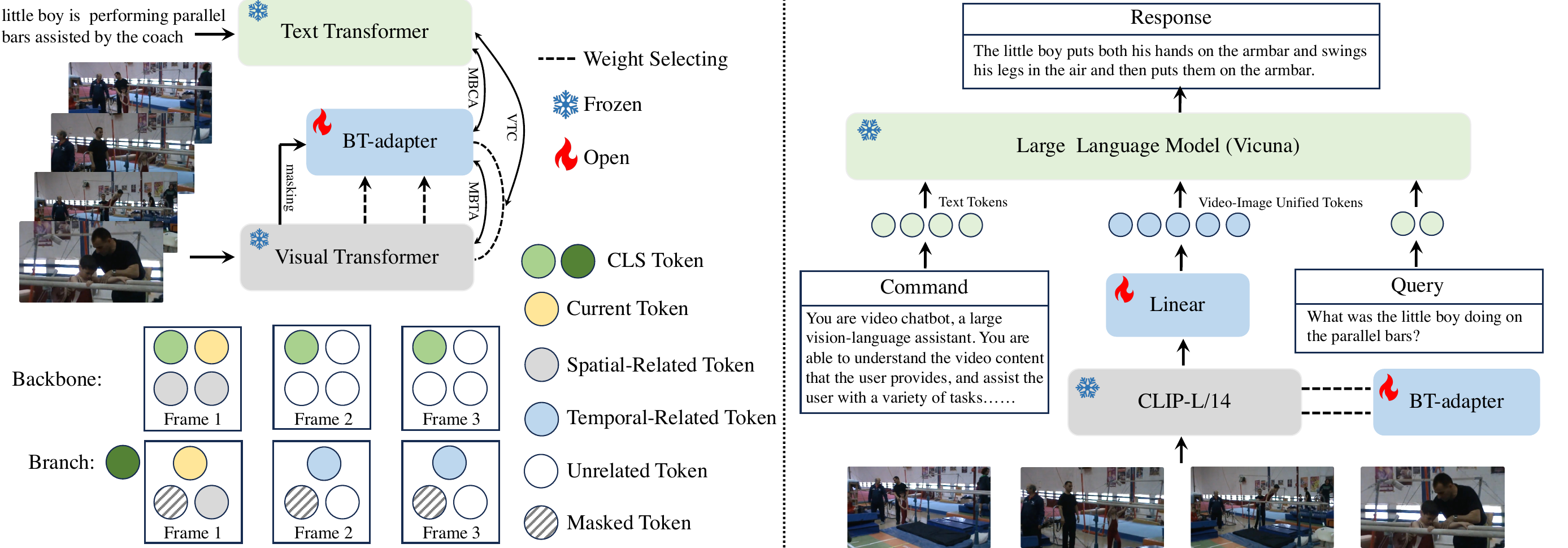} 
    \caption{The overview of our model. The left side shows the model architecture and the data flow during pretraining. The right side depicts the pipeline of video conversation.}
    \label{image_method}
    \vspace{-1.5em}
\end{figure*}

In this section, we elaborate on how to efficiently enhance pretrained image-language models (\textit{e.g.,}, CLIP \citep{radford2021learning} and LLaVA \citep{liu2023visual}) with temporal modeling capabilities while preserving the pretrained knowledge. The main framework is depicted in Fig. \ref{image_method} . We will introduce the model architectures of BT-Adapter in Sec. \ref{method_arch}, the exploration of temporal modeling in video dialogue models in Sec. \ref{method_temp}, and the training strategy and objectives in Sec. \ref{method_training}.


\subsection{Branching Temporal Adapter} \label{method_arch}

As the pioneer of contrastive image-text pretraining, CLIP \citep{radford2021learning} has founded widespread application in various domains. Meanwhile, fundamental image dialogue models like BLIP-2 \citep{li2023blip} and LLaVA \citep{liu2023visual} all employ CLIP as visual encoder. Hence, without losing the generalizability, we focus on the adaption of CLIP.

\vspace{0.2em}

\noindent \textbf{Model Architecture.} 
To enable video input, within CLIP, we treat each frame as an individual image. Given a video with T frames, we divide each frame into N non-overlapping patches, represented as $V = \{V_i\}_{i=1}^T$ and $V_i = \{v_{i,j}\}_{j=0}^N$, where $v_{i,0}$ denotes the [CLS] token. Then, tokens in each frame are added with spatial position and fed independently into the $l^{th}$ CLIP layers:
\begin{equation}
\begin{aligned}
    V_i^{'(l-1)} = \mathrm{S\_attn}(\mathrm{LN}(V_i^{(l-1)})) + V_i^{(l-1)}, \\
    V_i^{(l)} = \mathrm{FFN}(\mathrm{LN}(V_i^{'(l-1)})) + V_i^{'(l-1)}, \label{s_att}
\end{aligned}
\end{equation}
where $\mathrm{S\_attn}$, $\mathrm{LN}$ and $\mathrm{FFN}$ denote spatial attention, layer normalization and feed-forward network.

In contrast to the plug-style adapter that consists of multiple independent modules, BT-Adapter is a continuous network operating as a branch alongside the main backbone. Inside BT-Adapter, we adopt divided space-time attention \citep{bertasius2021space}. Given all video tokens, we first gather patch tokens in the same position across different frames, obtaining $\hat{V} = \{\hat{V_j}\}_{j=1}^N$ and $\hat{V_j} = \{\hat{v}_{i,j}\}_{i=1}^T$. Then, tokens in each position are fed individually into the $l^{th}$ temporal layers:
\begin{equation}
    \hat{V}_j^{'(l-1)} = \mathrm{W_t} \cdot  \mathrm{T\_attn}(\mathrm{LN}(\hat{V}_j^{(l-1)})) + \hat{V}_j^{(l-1)}, \\  
\end{equation}
where $\mathrm{T\_attn}$ is the self-attention operated on the T dimension. $\mathrm{W_t}$ is an additional zero-initialized linear projection, which keeps the training stable during the adaption. Then, the position-wise tokens are reshaped into frame-wise tokens $\{\hat{V'}_i\}_{i=1}^T$, and fed into the spatial layer, following the same procedure as CLIP layer in Eq. \ref{s_att}.


\noindent \textbf{Backbone-Branch Interaction.} 
Unlike traditional adapters, which are added to every pretrained layer, our branching adapter has significantly fewer layers compared to the main backbone. Assuming we have K layers in the branch, BT-Adapter takes the output of the last K+1 CLIP layers as input, each layer on both sides corresponding pairwise. To construct the input of the first BT-Adapter layer from the $k^{th}$ CLIP layer, we first develop a new learnable video [CLS] token $\hat{v}_{0,0}^{(0)}$ to represent the entire video. Then, we concatenate $\hat{v}_{0,0}^{(0)}$ with $V^{(k)}$ and update the patch embeddings with frozen spatial positional embeddings and learnable temporal positional embeddings:
\begin{equation}
    \hat{v}_{i,j}^{(0)} = v_{i,j}^{(k)} + \mathrm{P}_i^t + \mathrm{P}_j^s, \\ 
\end{equation}
where $P^s$ is the spatial position embedding shared with CLIP while $P^t$ is the temporal position embedding. For any other layer of BT-Adapter, we construct its input $\hat{V}^{(l)}$ from the previous branching layer and the CLIP layer at the same level with weight selecting as follows: 
\begin{equation}
    \hat{v}_{i,j}^{(l)} = \mathrm{Sigmoid(W_b)} \cdot \hat{v}_{i,j}^{(l-1)} + 
    \\ (1-\mathrm{Sigmoid(W_b)}) \cdot v_{i,j}^{(k+l-1)}, \\  
\end{equation}
where $\mathrm{W_b}$ is the learnable selective weight with zero initialization. At the final stage, the output of the branch and CLIP are combined for the out-of-the-box representation:
\begin{equation}
\begin{split}
    v = \mathrm{W_{v\_proj}}(\mathrm{LN}(\mathrm{Sigmoid(W_b)} \cdot \hat{v}_{0,0}^{(-1)} \\ + (1-\mathrm{Sigmoid(W_b)}) \cdot \frac{1}{T} \sum_{i=1}^{T} v_{i,0}^{-1} )), \label{eq_combine}
\end{split}
\end{equation}
where $\mathrm{W_{v\_proj}}$ is the frozen weight in CLIP projecting the visual embedding into joint visual-text feature space. With CLIP kept frozen, the backbone encodes low-level spatial patterns and high-level aligned features, while the branch focuses on modeling temporal dependencies. In this way, we leverage strong temporal modeling capacities while preserving the pretrained knowledge intact.

\subsection{Temporal Modeling for Video Conversation} \label{method_temp}
In this section, we conduct an empirical study to explore potential temporal modeling strategies for video dialogue models, demonstrating the advantages of our approach. We argue that an ideal temporal modeling approach for video-LLM models should meet several criteria: it should be parameter-efficient (P-E), as image dialogue models are already quite large; it should be multimodal-friendly (M-F), preserving as much multimodal alignment knowledge as possible; and it should be temporal-sensitive (T-S), delivering strong performance in time-sensitive scenarios. To measure the ``multimodal-friendly" and ``temporal-sensitive", we employ the ``Correctness of Information" and ``Temporal Understanding" metrics from the VideoChatGPT benchmark \citep{maaz2023video} respectively. ``parameter-efficient'' is simply decided by whether the CLIP can be frozen. Next, we will explain how we implement various temporal modeling methods for video conversation.

\noindent \textbf{Spatial-Temporal Pooling.}
Following \citet{maaz2023video}, we do not integrate any module in CLIP and encode each frame independently. Then, all patch tokens $V \in \mathbf{R}^{T \times N}$ are pooled along the time ($T$) and spatial ($N$) dimensions, resulting in a total of $T+N$ tokens. These tokens are subsequently input into the LLM, which has much fewer tokens compared to joint or separate temporal modeling.

\noindent \textbf{Joint-ST modeling.} 
Following \citet{xue2022clip}, we incorporate spatiotemporal positional embeddings to 2D patches and feed all $T*N$ video tokens into CLIP and LLM simultaneously. In this way, tokens in any position or frame can attend to each other, providing a straightforward yet effective method for modeling temporal dependencies. To make training feasible, we utilize the FSDP \citep{zhao2023pytorch} to facilitate parameter and gradient sharing between GPUs.

\begin{table}[]
    \setlength{\abovecaptionskip}{0.cm}
    \setlength{\belowcaptionskip}{-0.3cm}
    \begin{center}
      \caption{The finetuning results of different temporal modeling on video conversation. * means CLIP is frozen.} 
      \label{method_study}
      \renewcommand\tabcolsep{9pt}
      \resizebox{1.\linewidth}{!}{
      \begin{tabular}{l|ccc|c|c} 
      \toprule
      \multirow{2}{*}{Model} & \multirow{2}{*}{P-E} & \multirow{2}{*}{M-F} & \multirow{2}{*}{T-S} & Correctness of & Temporal \\  
     &  &  &  & Information & Understanding \\ \midrule
     Baseline* &  &  & & 2.38 & 1.93 \\ 
     ST Pooling* & $\checkmark$ & $\checkmark$ & & 2.40 & 1.98 \\ 
     Joint-ST & & & $\checkmark$ & 1.92 & 2.11 \\ 
     Separate-ST & & & $\checkmark$ & 2.10 & 2.13 \\ 
     Separate-ST* & $\checkmark$ & & & 2.29 & 2.01 \\ 
     BT-Adapter* & $\checkmark$ & $\checkmark$ & $\checkmark$ & \textbf{2.55} & \textbf{2.26} \\ 
      \bottomrule
    \end{tabular} }
    \end{center}
     \vspace{-2em}
  \end{table}
  
\noindent \textbf{Separate-ST modeling.}
Unlike joint-ST modeling, separate-ST modeling retains and leverages CLIP's spatial layer. All $T*N$ patch tokens are fed in LLM. Following \citet{zeng2023tvtsv2}, we insert temporal attention before each CLIP layer and add temporal position embeddings to the input. Regarding the backbone, separate-ST modeling offers flexibility, allowing us to experiment with both frozen and unfrozen CLIP layers.

\noindent \textbf{Branch Temporal Adapter.}
We directly equip CLIP-L/14 with BT-Adapter in Sec. \ref{method_arch} to achieve video encoder. In LLaVA, LLM takes the output from the second-to-last layer of the visual encoder as its input. Hence, we also take the second-to-last output from both CLIP and BT-Adapter and combine them with learnable balance weight. Finally, combined patch tokens $V \in \mathbf{R}^{T \times N}$ are pooled along the time ($T$) dimension, resulting in only $N$ tokens for inputting into LLM, which is the least among all methods.

\noindent \textbf{Results.} We replace CLIP-L/14 in LLaVA with the implemented video encoders and then proceed with video instruction tuning. In all methods, we keep LLM frozen while opening the linear projection between the visual encoder and LLM. No pretraining on video-text datasets is included for fair comparison.
Results are presented in Table \ref{method_study}. It can be observed that methods with a frozen CLIP tend to be parameter-efficient but perform poorly in terms of temporal modeling. On the other hand, methods with good temporal understanding suffer from high computation costs and knowledge loss. In contrast, our BT-Adapter achieves a balance by being parameter-efficient, multimodal-friendly, and temporal-sensitive simultaneously. The results reveal that BT-Adapter has a clear advantage over other temporal modeling methods for video conversation.

\subsection{Pretraining with Asymmetric Masking} \label{method_training}
As proved by previous studies \citep{wang2022internvideo, xue2022clip}, CLIP-based video encoder can harvest stronger performance on downstream video tasks after being post-pretrained on large-scale video-text data. Hence, we also involved our BT-Adapter with video-text pretraining. 
However, the expensive computation cost has always been a bottleneck when scaling up the training. Inspired by the recent success of masked modeling in visual-language pretraining \citep{li2023scaling,tong2022videomae}, we develop a unique asymmetric masking strategy for BT-Adapter. Specifically, we maintain tokens in the frozen CLIP unchanged while applying a tubular mask to the tokens in BT-Adapter. This mask randomly masks a certain percentage ($\rho\%$) of patch tokens in the same position across different frames. This approach allows us to maximize the retention of pretraining knowledge from CLIP while reducing spatial-temporal redundancy. Thanks to the asymmetric mask, we can maintain a high mask ratio ($\rho \ge 70$) without compromising performance, leading to a reduction of at least half of the computational budget. As a result, with the assistance of frozen backbone and token masking, we can accomplish the costy video-text pretraining in just a few hours.
Furthermore, based on asymmetric masking, we have devised two tailor-made training objectives for the branching temporal structure, in addition to the Video-Text Contrastive.

\noindent \textbf{Video-Text Contrastive (VTC).}  VTC is the most widely-used basic objective for cross-modal alignment. Given the global video feature $v$ in Eq. \ref{eq_combine} and global text feature $t$, we formulate $\mathcal{L}_{VTC}$ as:
\begin{equation}
\begin{split}
    & \mathcal{L}_{nce}(x,y) = - \frac{1}{B} \sum_{m=1}^B \mathrm{log} \frac{\mathrm{exp}(\tau x_{m} \cdot y_{n})}{\sum_{n=1}^B \mathrm{exp}(\tau x_{m} \cdot y_{n})}, \\
    & \mathcal{L}_{VTC} = \mathcal{L}_{nce}(v,t) + \mathcal{L}_{nce}(t,v),
\end{split}
\end{equation}
 where $B$ is the batch size and $m,n$ is the index in a batch and $\tau$ is the temperature scale.

\begin{table*}[t]
\setlength{\tabcolsep}{3pt}
\centering
\caption{The results of video conversation on video-based generative
performance benchmarking. FT and ZS mean with and without video instruction tuning respectively.}
\vspace{-0.8em}
\resizebox{0.92\textwidth}{!}{
\begin{tabular}{l|c|c|c|c|c|c}
\toprule
\multicolumn{1}{c|}{\multirow{2}{*}{Method}} & Temporal & Correctness of & Detail & Contextual & \multirow{2}{*}{Consistency} & \multirow{2}{*}{Mean Score} \\
 & Understanding & Information &  Orientation & Understanding &  & \\ \midrule
 VideoLLaMA \cite{zhang2023video} & 1.82 & 1.96 & 2.18 & 2.16 & 1.79 &  1.98 \\
 LLaMA-Adapter \cite{gao2023llama} & 1.98 & 2.03 & 2.32 & 2.30 & 2.15 & 2.16 \\
 VideoChat \cite{li2023videochat} & 1.94 & 2.23 & 2.50 & 2.53 & 2.24 & 2.29 \\
 VideoChatGPT \cite{maaz2023video} & 1.98 & 2.40 & 2.52 & 2.62 & 2.37 & 2.38 \\
 \rowcolor[RGB]{207,234,241} BT-Adapter-LLaVA (ZS) & 2.13 & 2.16 & 2.46 & 2.89 & 2.20 & 2.46 \\
 \rowcolor[RGB]{207,234,241} BT-Adapter-LLaVA (FT) & \textbf{2.34} & \textbf{2.68} & \textbf{2.69} & \textbf{3.27} & \textbf{2.46} & \textbf{2.69} \\

\bottomrule
\end{tabular}
}
\label{chat_generate}
\vspace{-0.5em}
\end{table*}

\begin{table*}[]
  \begin{minipage}[b]{0.54\textwidth}
    \centering
    \caption{The results of video conversation zero-shot question-answering. FT and ZS mean with and without instruction tuning respectively.}
\vspace{-0.8em}
\resizebox{1.\textwidth}{!}{
\begin{tabular}{l|cc|cc|cc}
\toprule
\multirow{2}{*}{Method}  & \multicolumn{2}{c|}{MSVD-QA} & \multicolumn{2}{c|}{MSRVTT-QA} & \multicolumn{2}{c}{ActivityNet-QA} \\
 & Acc & Score & Acc & Score & Acc & Score \\ \hline
 VideoLLaMA  & 51.6 & 2.5 & 29.6 & 1.8 & 12.4 & 1.1 \\ 
 LLaMA-Adapter & 54.9 & 3.1 &43.8 &2.7 & 34.2 & 2.7 \\ 
 VideoChat  & 56.3 & 2.8 & 45.0 & 2.5 & 26.5 & 2.2 \\ 
 VideoChatGPT  & 64.9 & 3.3 & 49.3 & 2.8 & 35.2 & 2.7 \\ 
 \rowcolor[RGB]{207,234,241} Ours (ZS) & 67.0 & 3.6 & 51.2 & 2.9 & \textbf{46.1} & \textbf{3.2} \\ 
 \rowcolor[RGB]{207,234,241} Ours (FT) & \textbf{67.5} & \textbf{3.7} & \textbf{57.0} & \textbf{3.2} & 45.7 & \textbf{3.2} \\ 

\bottomrule
\end{tabular}}
\label{chat_QA}
\vspace{-0.5em}
  \end{minipage}%
   \hfill
  \begin{minipage}[b]{0.42\textwidth}
    \raggedleft
    \caption{The results of zero-shot video conversation on different image-centric dialogue models.}
\vspace{-0.8em}
\resizebox{1.\textwidth}{!}{
\begin{tabular}{l|ccc}
\toprule
Method  & Temporal & Correctness \\ \hline
 LLaVA-Vicuna  & 1.78 & 2.06 \\ 
 \rowcolor[RGB]{207,234,241} +BT-Adapter & 2.13 & 2.16 \\  \hline
 MiniGPT4-Vicuna   & 1.88 & 2.48 \\ 
 \rowcolor[RGB]{207,234,241} +BT-Adapter & 2.56 & 2.71 \\ \hline
 InstructBLIP-Vicuna  & 2.03 & 2.90 \\ 
\rowcolor[RGB]{207,234,241} +BT-Adapter & 2.38 & 2.93 \\  

\bottomrule
\end{tabular}}
\label{chat_zs}
\vspace{-0.5em}
  \end{minipage}
\end{table*}

\noindent \textbf{Masked Branching Token Alignment (MBTA).} 
Masked video modeling has been proven beneficial for spatial-temporal representation learning \citep{tong2022videomae}, but pixel reconstruction in \citet{tong2022videomae} is computationally expensive. In our approach, we have a frozen CLIP and the masked BT-Adapter, where the unmasked CLIP is naturally a teacher for the masked branch. Hence, we can conduct the token alignment in an end-to-end pretraining without extra forward propagation. 
Specifically, we compute the mean squared error (MSE) between the unmasked tokens in BT-Adapter and the corresponding tokens in CLIP in the last layer, formulating $\mathcal{L}_{MBTA}$ as:
\begin{equation}
\begin{split}
    \mathcal{L}_{MBTA} & = \frac{1}{(1-\rho) N T} \sum_{i=1}^{T} \sum_{j=1}^{N} ( \mathrm{W_{v\_proj}} \cdot v_{i,j}^{-1} \\ 
    & - \mathrm{W'_{v\_proj}} \cdot \hat{v}_{i,j}^{-1})^2, ~~(i,j) \notin M,
\end{split}
 \end{equation}
 where $M$ is the set of masked tokens, and $\mathrm{W'_{v\_proj}}$ is a freshly initialized linear projection distinct from $\mathrm{W_{v\_proj}}$. Utilizing MBTA, we can constrain the masked branch to reconstruct and align the semantic information present in CLIP, leading to better multimodal results.

 \noindent \textbf{Masked Branching Cross-Modal Alignment (MBCA).} 
In our model, the backbone and the branch handle different aspects: CLIP encodes static spatial features, while BT-Adapter captures dynamic information. Therefore, we additionally align the branching patch tokens with the text embedding to align the temporal-sensitive information, which is formulated as:
\begin{equation}
\begin{split}
    & \hat{v} = \frac{1}{(1-\rho) N T} \sum_{i=1}^{T} \sum_{j=1}^{N} \mathrm{W_{v\_proj}} \cdot \hat{v}_{i,j}^{-1}, \\
    & \mathcal{L}_{MBCA} = \mathcal{L}_{nce}(\hat{v},t) + \mathcal{L}_{nce}(t,\hat{v}), ~~(i,j) \notin M.
\end{split}
 \end{equation}
Different from $\mathcal{L}_{VTC}$, we utilize the patch tokens for the alignment. This choice is made because the information is highly centralized in the CLS token for CLIP, whereas downstream applications like multimodal conversation or generation typically rely on the patch tokens as input.

\section{Experiments} \label{exp}

\subsection{Experiment Settings} 
\noindent \textbf{Tasks and Datasets.}
We have evaluated our BT-Adapter on two main aspects: traditional video tasks and video-centric dialogue. To begin with, we pretrained the BT-Adapter on WebVid-2M \citep{bain2021frozen}. For traditional video tasks, we consider zero-shot text-to-video retrieval and video recognition, covering seven benchmarks: (a)MSR-VTT \citep{xu2016msr}. (b)DiDeMo \citep{anne2017localizing}. (c)LSMDC \citep{rohrbach2017movie}. (d)ActivityNet \citep{caba2015activitynet}. (e)Kinetic-400 \citep{carreira2017quo}. (f)HMDB-51 \citep{kuehne2011hmdb}. (g)UCF-101 \citep{soomro2012ucf101}. We used Recall@K (R@K) for text-to-video retrieval and Top-K Accuracy (A@K) for video recognition as the evaluation metrics.

For video dialogue, we conducted evaluations for the zero-shot video conversation (without instruction tuning) and instruction-tuned video conversation. The VideoChatGPT-100K \citep{maaz2023video} is employed for supervised instruction tuning. To assess the quality of the responses quantitatively, we employ the VideoChatGPT benchmark, which consists of five metrics for video-based generative performance benchmarking and six metrics for zero-shot question-answer evaluation. Additional details and settings of each dataset can be found in Sec. \ref{appendix_dataset} in the Appendix.

\noindent \textbf{Implementation Details.}
We employ openai-CLIP-L/14 and EVA-CLIP-G \cite{sun2023eva} as the video backbone. For all downstream tasks, we use openai-CLIP-L/14. We adopt 4 layers of BT-Adapter in default. During pretraining, the masking ratio $\rho$ is set as 70\%. The temperature scale $\tau$ is fixed as 0.01 for contrastive loss. The weight of the three losses is $1:1:1$. It takes 3 hours to train one epoch on WebVid on 8 V100-32G GPUs. For video conversation, we implement the instruction tuning based on BT-Adapter-LLaVA-7B. Unless specifically noted, we use
BT-Adapter-LLaVa in all video dialogue tasks. InstructBLIP and MiniGPT4 are also included for zero-shot evaluation. All conversation models are based on Vicuna1.0-7B \cite{zheng2023judging}. During instruction tuning, we update the linear projection and BT-Adapter, while keeping the rest architecture frozen. It takes 3 hours to train three epochs on 8 A100 40GB GPUs. More training details are listed in Appendix.

\begin{table*}[t]
\setlength{\tabcolsep}{3pt}
\centering
\caption{The zero-shot results of text-to-video retrieval on MSR-VTT, DiDemo, LSMDC, and ActivityNet.}
\vspace{-0.8em}
\resizebox{0.95\textwidth}{!}{
\begin{tabular}{l|c|ccc|ccc|ccc|ccc}
\toprule
\multirow{2}{*}{Method} & Pretraining & \multicolumn{3}{c|}{MSR-VTT} & \multicolumn{3}{c|}{DiDeMo} & \multicolumn{3}{c|}{LSMDC} & \multicolumn{3}{c}{ActivityNet}   \\

                                              & Scale  & R@1   & R@5   & R@10  & R@1   & R@5   & R@10  & R@1  & R@5  & R@10   & R@1  & R@5  & R@10   \\
\hline
\textit{Non-CLIP models} & & & & & & & & & & & & & \\
Frozen \citep{bain2021frozen}   & 5M  & 24.7  & 46.9  & 57.2  & 	21.1  & 46.0  & 56.2  & - & - & - & - & - & - \\
Clover \citep{huang2023clover}   & 5M  & 26.4  & 49.5  & 60.0  & 29.5  & 55.2  & 66.3  & 17.4 & 29.2 & 38.2 & - & - & - \\
OmniVL \citep{wang2022omnivl}     & 14M  &  34.6 & 58.4 & 66.6  & 33.3 & 58.7 & 68.5  & - & - & - & - & - & - \\
HiTeA \citep{ye2022hitea}   & 5M  &  29.9 & 54.2 & 62.9  & 36.1 & 60.1 & 70.3  & 15.5 & 31.1 & 39.8 & - & - & - \\
Singularity \citep{lei2022revealing}   & 17M  &  34.0 & 56.7 & 66.7  & \textbf{37.1} & 61.7 & 69.9  & - & - & - & 30.6 & 55.6 &66.9 \\
VideoCoCa \citep{yan2022video}   & 100M  &  34.3 & 57.8 & 67.0  & - & - & -  & - & - & - & 34.5 & 63.2 & 76.6 \\ \hline
\textit{CLIP-L/14} & & & & & & & & & & & & & \\
CLIP \citep{radford2021learning}     & -  &  35.4 & 58.8 & 68.1  & 30.3 & 54.9 & 65.4  & 17.0 & 31.8 & 40.3 & 28.8 & 57.6 & 71.8 \\
ImageBind \citep{girdhar2023imagebind}   & -  &  36.8 & 61.8 & 70.0  & - & - & -  & - & - & - & - & - & - \\
InternVideo \citep{wang2022internvideo}     & 12.8M  & 40.7  & -  & -  & 31.5  & -  & -  & 17.6 & - & - & 30.7 & - & - \\
TVTSv2 \citep{zeng2023tvtsv2}     & 8.5M  & 38.2  & 62.4  & 73.2  & 34.6  & \textbf{61.9}  & 71.5  & 17.3 & 32.5 & 41.4 & - & - & - \\
UMT-L \citep{li2023unmasked}     & 5M  & 33.3  & 58.1  & 66.7  & 34.0  & 60.4  & 68.7  & \textbf{20.0} & \textbf{37.2} & 43.7 & 31.9 & 60.2 & 72.0 \\
\rowcolor[RGB]{207,234,241}  BT-Adapter      & 2M  & \textbf{40.9}  & \textbf{64.7}  & \textbf{73.5}  & 35.6  & \textbf{61.9}  & \textbf{72.6}  & 19.5 & 35.9 & \textbf{45.0} & \textbf{37.0}  & \textbf{66.7}  & \textbf{78.9}\\

\bottomrule
\end{tabular}}
\label{zs_retrieval}
\vspace{-0.5em}
\end{table*}

\begin{table*}[]
  \begin{minipage}[b]{0.58\textwidth}
    \centering
    \caption{Ablation study on the structures of BT-Adapter. We report the results on zero-shot R@1 of MSRVTT and DiDemo and zero-shot video conversation.}
\vspace{-0.8em}
\resizebox{1.\textwidth}{!}{
\begin{tabular}{l|ccc}
\toprule
 Model & MSRVTT  & DiDemo & Temporal \\
 \hline
 CLIP (baseline) & 35.4 & 30.3 & 1.78 \\ 
 +4 layer separate-ST & 35.7 & 31.0 & 1.81 \\
 +branch modeling & 37.4 & 32.9 & 1.97 \\
 +backbone-branch interaction & \textbf{38.5} & \textbf{33.9} & \textbf{2.06} \\ 

\bottomrule
\end{tabular}}
\label{structure}
\vspace{-0.5em}
  \end{minipage}%
   \hfill
  \begin{minipage}[b]{0.38\textwidth}
    \raggedleft
    \caption{Ablation study on the training objectives. We report the zero-shot R@1 of retrieval.}
\vspace{-0.8em}
\resizebox{1.\textwidth}{!}{
\begin{tabular}{cc|cc}
\toprule
MBTA  & MBCA & MSRVTT & DiDemo \\ \hline
 &  & 38.5 & 33.9 \\ 
 $\checkmark$ &  & 39.3 & 34.3 \\ 
   & $\checkmark$ & 40.1 & 34.9 \\ 
  $\checkmark$  & $\checkmark$ & \textbf{40.9} & \textbf{35.6} \\ 

\bottomrule
\end{tabular}}
\label{objective}
\vspace{-0.5em}
  \end{minipage}
\end{table*}

\subsection{Quantitative evaluation} 
\textbf{Video Dialogue.} 
Thanks to the benchmarks introduced by \citet{maaz2023video}, we can quantitatively compare the performance of various video conversation models. The results are presented in Table \ref{chat_generate} and Table \ref{chat_QA}, where we compare BT-Adapter with all existing video-centric dialogue models, including VideoLLaMA \citep{zhang2023video}, LLaMA-Adapter \citep{gao2023llama}, VideoChat \citep{li2023videochat}, and VideoChatGPT \citep{maaz2023video}. It is observed that our zero-shot model, even without any instruction tuning, outperforms methods that require instruction tuning on average. When fine-tuned using video instruction data, the superiority of our approach becomes even more pronounced. These results underscore the effectiveness of the BT-Adapter as a superior method for temporal modeling in video conversation models compared to existing approaches. Notably, our BT-Adapter exhibits a significantly larger performance margin over other methods on ActivityNet in both Table \ref{chat_QA} and \ref{zs_retrieval}, highlighting its particular strength in handling long video sequences. Moreover, we have integrated the BT-Adapter with various pretrained image conversation models. As illustrated in Table \ref{chat_zs}, pretrained BT-Adapter yields consistent advancement on all image-language chatbots without extra instruction tuning, underscoring the broad applicability of our method.

\vspace{-0.1em}

\noindent \textbf{Traditional Video Tasks.}
The results of zero-shot text-to-video retrieval are presented in Table \ref{zs_retrieval}. Compared to the previous SOTAs, BT-Adapter achieves competitive performance across all datasets, obtaining the best results on most metrics. For instance, in comparison to UMT and Singularity, although there is a slight lag in terms of R@1 on DiDeMo and LSMDC respectively, we have surpassed them by more than 5\% on MSRVTT and ActivityNet. Furthermore, we achieve superior results with significantly fewer pretraining scales and GPU hours than all the mentioned methods. For example, we use 130$\times$ fewer GPU hours than UMT, 560$\times$ fewer than TVTSv2, and 2687$\times$ fewer than InterVideo. The results of zero-shot action recognition are posted in Table \ref{zs_action} in Appendix.

\vspace{-0.2em}

\begin{figure*}[htbp] 
    \centering
    \begin{minipage}{0.48\textwidth}
      \centering
      \includegraphics[width=1\textwidth]{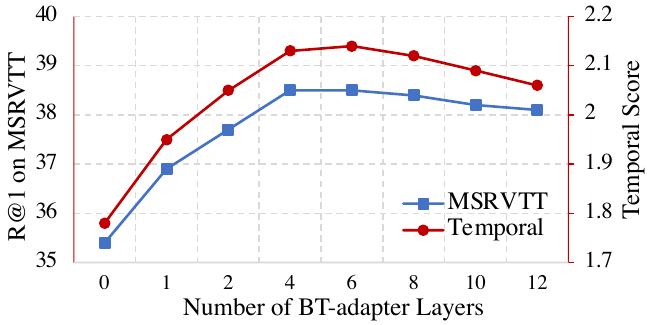} 
      \vspace{-1.7em}
      \caption{Ablation study on the number of BT-Adapter layers.}
      \label{layernum}
      \vspace{-0.8em}
    \end{minipage}
    \hfill
    \begin{minipage}{0.48\textwidth}
      \centering
      \includegraphics[width=1\textwidth]{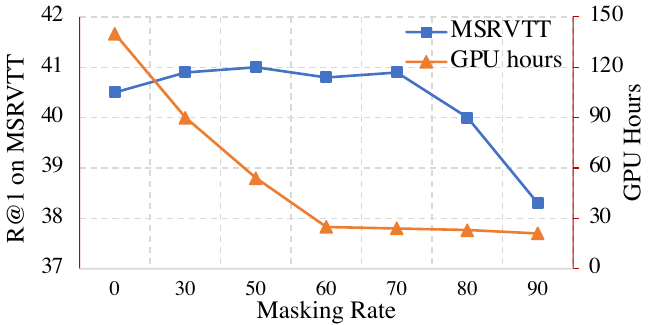} 
      \caption{Ablation study on the masking rate during pretraining.}
      \label{maskrate}
      \vspace{-0.8em}
    \end{minipage}
\end{figure*}

\begin{figure*}[t] 
    \centering
    \includegraphics[width=1\textwidth]{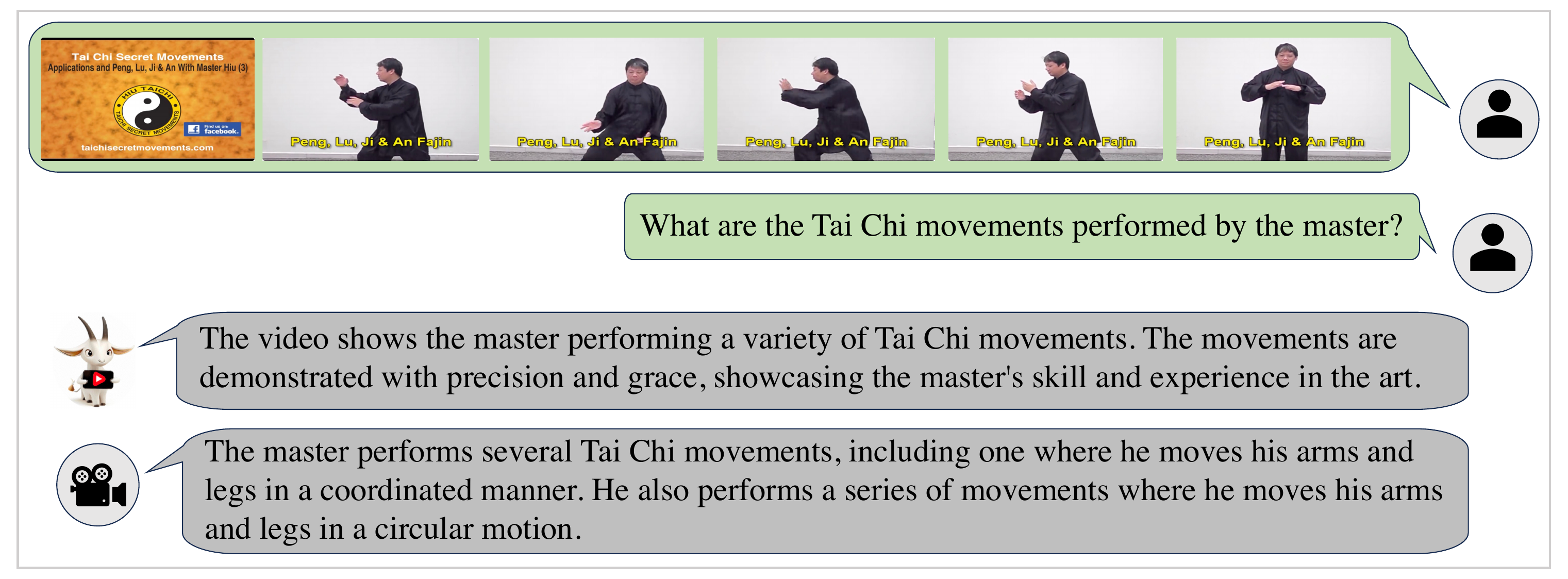} 
    \captionsetup{font={footnotesize}}
    \vspace{-2.5em}
    \caption{A qualitative result of video conversation. We present the answers from VideoChatGPT \citep{maaz2023video} (upper) and our BT-Adapter-LLaVA (down).}
    \label{visualize1}
    \vspace{-1em}
\end{figure*}

\subsection{Ablation Study} 
\textbf{Model Structure.} 
To investigate the components within the BT-Adapter, we initiated our study by exclusively employing VTC for pretraining and subsequently evaluating the zero-shot performance across retrieval tasks and video conversation, as presented in Table \ref{structure}. Initially, we explored the implementation of separate-ST modeling within the last 4 layers. However, the resulting improvements were marginal. Subsequently, we transitioned the separate-ST network into the branch, where we witnessed notable progress. This shift underscores the efficacy of branching modeling. In the final step, we studied the interaction module, \textit{i.e.,} multi-level selective combination between backbone and branch. This addition led to further enhancements across all benchmark datasets. More fine-grained ablations on model components can be found in Sec. \ref{appendix_ablation} in appendix.

\noindent \textbf{Training Objectives.}
We investigated the influence of two innovative training objectives in Table \ref{objective}, where all experiments were conducted using BT-Adapter with a mask rate of 70\%. Our observations reveal that both MBTA and MBCA yield improvements in downstream results, with MBCA demonstrating more substantial progress. This outcome suggests that individually aligning temporal modules and textual outputs effectively mitigates the limitations inherent in pretrained image-text models. Moreover, the combination of both objectives results in further advancements.

\noindent \textbf{Number of Branching Layers and Masking Rate.} 
The number of branching layers and the masking rate in our model inherently involve a trade-off between computational resources and performance. To determine the optimal settings, we conducted ablation experiments. Firstly, in Fig. \ref{layernum}, we present the results of zero-shot retrieval and conversation as we vary the number of layers. Notably, we observed that the performance improvement plateaus at around 4-6 layers. Secondly, in Fig. \ref{maskrate}, we provide results for zero-shot retrieval and the corresponding convergence GPU time across different masking rates. Our findings indicate that the results remain stable when the masking rate is set at or below 70\%, while higher masking rates significantly lower the training time. Therefore, we have chosen to maintain a masking rate of 70\%.

\subsection{Qualitative Results} 
In Fig. \ref{visualize1}, we present a qualitative example of video conversation. Unlike the general description from VideoChatGPT, our model provides an informative and accurate response to a video-related question, highlighting the effectiveness of the BT-Adapter in video comprehension. Additional examples of video dialogues covering various aspects can be found in the Appendix.

\section{Conclusion}
This paper presents a novel approach to achieve parameter-efficient yet effective image-to-video adaptation and video conversation. Our proposed solution, named Branching Temporal Adapter (BT-Adapter), is a branching separate-ST network for temporal modeling. Building upon the BT-Adapter, we introduce an asymmetric masking technique along with two novel training objectives. Extensive experiments demonstrate the superiority of our model over other temporal modeling methods for video conversation. By seamlessly integrating the BT-Adapter with any pretrained image conversation model, we achieve video dialogues without necessitating manual video instruction tuning. With significantly lower computation costs, we attain state-of-the-art results across two zero-shot video tasks and video conversations.

{\footnotesize{
\noindent \textbf{Acknowledgements.} 
    This work was supported by National Natural Science Foundation of China (No. 62172021).}}

{
    \small
    \bibliographystyle{ieeenat_fullname}
    \bibliography{main}
}

\clearpage
\setcounter{page}{1}
\maketitlesupplementary

\section{Datasets} \label{appendix_dataset}
\textbf{Test-to-Video Retrieval.}
The settings of the four zero-shot retrieval benchmarks are presented as follows: (1) \textbf{MSRVTT} \citep{xu2016msr}, a widely used video-text retrieval benchmark, comprises 10,000 YouTube videos, each accompanied by 20 captions. Our reported results are based on the 1K-A split, which consists of 9,000 training videos and 1,000 testing videos. For MSRVTT, we sample 12 frames for each video and set max token length as 12. (2) \textbf{DiDemo} \citep{anne2017localizing} comprises 10,611 videos gathered from Flickr, along with 40,000 sentences. To form queries, we concatenate all captions associated with a video. We use a frame number of 64 and a mask token length of 64, consistent with prior research. (3) \textbf{LSMDC} \citep{rohrbach2017movie} consists of 118,081 videos extracted from 202 movies. We configure it with a frame number of 12 and a maximum token length of 32. (4) \textbf{ActivityNet} \citep{caba2015activitynet} comprises 20,000 YouTube videos. To create queries, we concatenate all video descriptions into paragraphs. Our evaluation focuses on video-paragraph retrieval using the 'val1' split. We set the frame number and maximum token length to 64.

\textbf{Action Recognition.}
In all video recognition datasets, we refrain from utilizing templates such as ``a video of **'' and instead employ the tag itself as the textual query. The parameters for frame number and maximum token length are consistently configured at 16 and 12 respectively. The statistics pertaining to the three zero-shot action recognition benchmarks are provided below: (1) \textbf{Kinetics-400} \citep{carreira2017quo} is a widely recognized dataset for video action recognition. It comprises a substantial collection of 260,000 videos, each with an average duration of approximately 300 frames. The dataset encompasses a diverse set of 400 action classes. (2) \textbf{HMDB-51} \citep{kuehne2011hmdb} includes a total of 5,000 videos spanning 51 distinct action categories. The dataset is partitioned into training and test sets, with 3,500 videos allocated for training and 1,500 videos for testing. (3) \textbf{UCF-101} \citep{soomro2012ucf101} comprises a comprehensive collection of 13,000 videos, representing 101 unique action categories. Within this dataset, the training set consists of 9,500 videos, while the test set contains 3,500 videos.

\begin{table*}[!h]
\setlength{\tabcolsep}{3pt}
\centering
\caption{Default implementation details for pretraining and instruction tuning.}
\vspace{-0.8em}
\resizebox{0.9\textwidth}{!}{
\begin{tabular}{lcc}
\toprule
Task & Video-Text Pretraining & Video Instruction Tuning  \\
\hline
num. BT-Adapter layers & 4 & 3 \\
num. CLIP layers & 24 & 23 \\
optimizer & AdamW, $\beta = (0.9, 0.98)$ & AdamW, $\beta = (0.9, 0.998)$  \\
weight decay & 0.05 & 0.1 \\
learning rate & 2e-6 (for BT-Adapter) & 2e-5 (for BT-Adapter and linear projection) \\
fp16 & \XSolid & \Checkmark \\
batch size & 640 & 4 \\
augmentation & RandomResizedCrop & CenterCrop  \\
training source &  8 V100-32G & 4 A100-40G \\

\bottomrule
\end{tabular}}
\label{implementation}
\end{table*}

\textbf{Video-Text pretraining.} We adopt the \textbf{WebVid2M} \citep{bain2021frozen} for pretraining, laying the foundation for the BT-Adapter's video encoding capabilities. WebVid2M is a substantial video-text pretraining dataset composed of short videos paired with textual descriptions, sourced from stock footage sites. This dataset is characterized by its vast scale, encompassing approximately 2.5 million video-caption pairs and totaling 12,000 video hours. The videos within WebVid2M exhibit a rich diversity of content. During the pretraining, we configured the frame number and maximum token length to be 8 and 32 respectively.

\textbf{Video Conversation.} 
VideoChatGPT benchmark \citep{maaz2023video} s the first benchmark designed for the quantitative evaluation of video conversation models. It was collaboratively annotated by ChatGPT and human annotators using the ActivityNet dataset, resulting in a dataset containing 100k video-text instruction pairs. For the video-based text generation benchmark, a test set was curated based on ActivityNet, which included captions and associated question-answer pairs obtained from human annotations. The evaluation pipeline used the GPT-3.5 model and assessed the model's performance in various aspects, including Correctness of Information, Detail Orientation, Contextual Understanding, Temporal Understanding, and Consistency. The pipeline assigns a relative score to the generated predictions on a scale of 1 to 5 for each of these aspects. For zero-shot question-answer evaluation, three open-source video QA datasets were employed: MSRVTT-QA, MSVD-QA, and ActivityNet-QA. Also, GPT was used as the zero-shot evaluation assistor to assign relative scores on a scale of 1 to 5 for generated answers.

\section{Implementation Details}
All experiments were conducted using PyTorch \citep{paszke2019pytorch}. The pretraining and zero-shot inference processes were implemented based on mmaction2.0 \citep{2020mmaction2}. Our configuration settings are detailed in Table \ref{implementation}, with the exception of specific cases where alternate configurations were used. It is noteworthy that our data augmentation techniques are notably simpler in comparison to those employed by other methods.

\section{Zero-Shot Results on Action Recognition} \label{sec_zsaction}
The results of zero-shot video recognition are reported in Table \ref{zs_action}. Despite being pretrained solely on video-language datasets, BT-Adapter consistently contributes to the video-only task, achieving state-of-the-art zero-shot results among CLIP-based methods. Notably, even when compared to InternVideo, which employed self-supervised reconstruction during pretraining (proven to be more effective on single-modality tasks than contrastive learning), BT-Adapter still outperforms it, underscoring the effectiveness of BT-Adapter in video encoding and spatial-temporal modeling.

\begin{table*}{}{}
\setlength{\tabcolsep}{3pt}
    \begin{center}
      \caption{ The zero-shot results of video recognition on HMDB, UCF, and K400.} 
      
      \renewcommand\tabcolsep{9pt}
      \resizebox{0.72\linewidth}{!}{
       \begin{tabular}{l|cc|cc|cc}
        \toprule
        \multirow{2}{*}{Method}  & \multicolumn{2}{c|}{HMDB-51} & \multicolumn{2}{c|}{UCF-101} & \multicolumn{2}{c}{K400}     \\
        
                                                        & A@1   & A@5   & A@1   & A@5   & A@1   & A@5     \\
        \hline
        JigsawNet \citep{qian2022rethinking}       &  38.7 & - & 56.0  & - & 45.9 & - \\
        CLIP \citep{radford2021learning}       &  45.0 & 74.4 & 73.5  & 92.7 & 59.1 & 82.8 \\
        X-Florence \citep{ni2022expanding}       &  48.4 & - & 73.2  & - & - & - \\
        InternVideo \citep{wang2022internvideo}    & -  & -  & -  & -  & 64.2  & -   \\
        TVTSv2 \citep{zeng2023tvtsv2}      & 52.1  & -  & 78.0  & - & 59.6  & -   \\
        \rowcolor[RGB]{207,234,241}  BT-Adapter      & \textbf{54.6}  & \textbf{79.7}  & \textbf{79.1}  & \textbf{96.2}  & \textbf{64.3}  & \textbf{86.7}    \\
        \bottomrule
        \end{tabular} 
        \label{zs_action}
    }
    \end{center}
    \vspace{-1.5em}
    
\end{table*}

\begin{table*}[]
\setlength{\tabcolsep}{3pt}
    \begin{center}
    \vspace{-1em}
      \caption{The experimental comparison with closely related works.} 
      \vspace{-1em}
      \renewcommand\tabcolsep{9pt}
      \resizebox{0.8\linewidth}{!}{
       \begin{tabular}{l|c|c|c|c}
        \toprule
        Method & MSR-VTT R@1 & Correctness & Temporal & GPU Hours \\ \hline
        CLIP(baseline) & 35.4 & 2.06 & 1.78 & - \\
        ST-Adapter & 33.6 & 1.52 & 1.71 & 2.5 \\
        STAN(open) & 40.5 & 1.84 & 1.77 & 22 \\
        STAN(frozen) & 38.1 & 2.07 & 1.92 & 3 \\
        \rowcolor[RGB]{207,234,241} BT-Adapter & 40.9 & 2.16 & 2.13 & 3 \\
        \bottomrule
        \end{tabular} 
        \label{similar_exp}
    }
    \end{center}
    \vspace{-0.5em}
\end{table*}

\section{Experimental Comparison With Similar Methods} \label{similar_sec}
In this section, we conduct an experimental comparison between two closely related works, ST-Adapter \citep{pan2022st} and STAN \citep{liu2023revisiting}. ST-Adapter is also notably recognized as a parameter-efficient method for temporal modeling, while STAN also employs the branching temporal modeling strategy. We pretrain the three methods on MSRVTT for one epoch first, and the results of zero-shot performance on MSRVTT retrieval and video conversation are presented in Table \ref{similar_exp}. Initially, it is evident that ST-Adapter exhibits suboptimal results across all metrics. This outcome may be attributed to the fact that ST-Adapter is a single-modality temporal adapter, where the insertion of 3-D convolutions between transformer layers may lead to the rapid degradation of the pretrained multimodal knowledge. Next, we assess STAN under two conditions: with frozen CLIP and without. The results reveal that STAN, when used with an open CLIP, performs admirably in zero-shot retrieval tasks. However, it exhibits poorer outcomes in video conversation tasks, and it requires significantly longer pretraining hours. Conversely, when STAN is employed with a frozen CLIP, it shows improvements across all metrics, although it still falls short of BT-Adapter in all aspects. In contrast, BT-Adapter achieves both efficiency and effectiveness simultaneously, underscoring the superiority of our design over ST-Adapter and STAN in the context of zero-shot video encoding and video conversation.

\section{More Ablation Results} \label{appendix_ablation}
\textbf{Temporal Projection and Initialization.} We examine the appropriate way for instantiating the temporal projection. As demonstrated in Table \ref{more_ablation}(above), random initialization for the projection yields performance results similar to those obtained without projection. In contrast, zero initialization outperforms them by a significant margin. This suggests that building temporal reasoning capability from scratch, as opposed to random initialization, mitigates adverse effects on the well-established spatial prior. Consequently, zero initialization is better suited for knowledge transfer from images to videos. 
\textbf{Backbone-Branch Combination.} We further perform an ablation study to explore the most effective method for combining the output from the backbone and the branch, considering three approaches: direct addition, weighted selection, and concatenation with subsequent linear projection. As illustrated in Table \ref{more_ablation}(below), weighted selection yields the most favorable results. This observation suggests that different layers and samples require distinct degrees of information from the backbone and the branch.

\begin{table}
\setlength{\tabcolsep}{3pt}
    \begin{center}
    \vspace{-1em}
      \caption{\small Ablation Studies on Temporal Projection and Backbone-Branch Combination.} 
      \vspace{-1em}
      \renewcommand\tabcolsep{9pt}
      \resizebox{1.\linewidth}{!}{
       \begin{tabular}{l|cc}
        \toprule
        Method & MSR-VTT R@1 & DiDemo R@1  \\ \hline
        None Projection & 39.1 & 33.5 \\
        Random Initialization & 39.3 & 34.0 \\
        Zero Initialization & \textbf{40.9} & \textbf{35.6} \\ \hline \hline
        Addition & 40.0 & 34.4 \\
        Weighted Selection & \textbf{40.9} & \textbf{35.6} \\  
        Concatenation & 39.8 & 34.5 \\ 
        \bottomrule
        \end{tabular} 
        \label{more_ablation}
    }
    \end{center}
    \vspace{-0.5em}
\end{table}

\begin{figure}[htb] 
    \centering
    \includegraphics[width=1\linewidth]{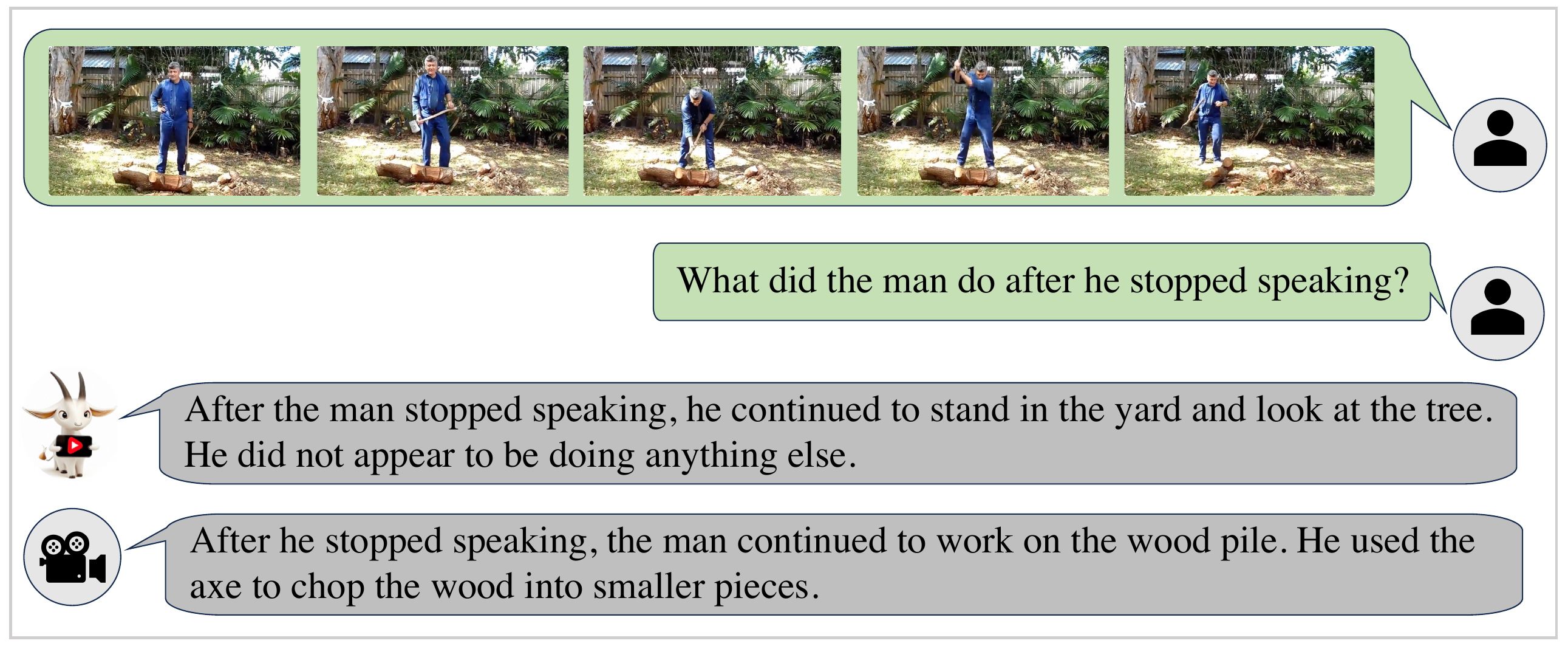} 
    \captionsetup{font={footnotesize}}
    \vspace{-1.8em}
    \caption{Qualitative results of video conversation in terms of the sequence of actions in the video. We present the answers from VideoChatGPT (upper) and our BT-Adapter-LLaVA (down).}
    \label{visualize11}
    \vspace{-0.5em}
\end{figure}

\begin{figure}[htb] 
    \centering
    \includegraphics[width=1\linewidth]{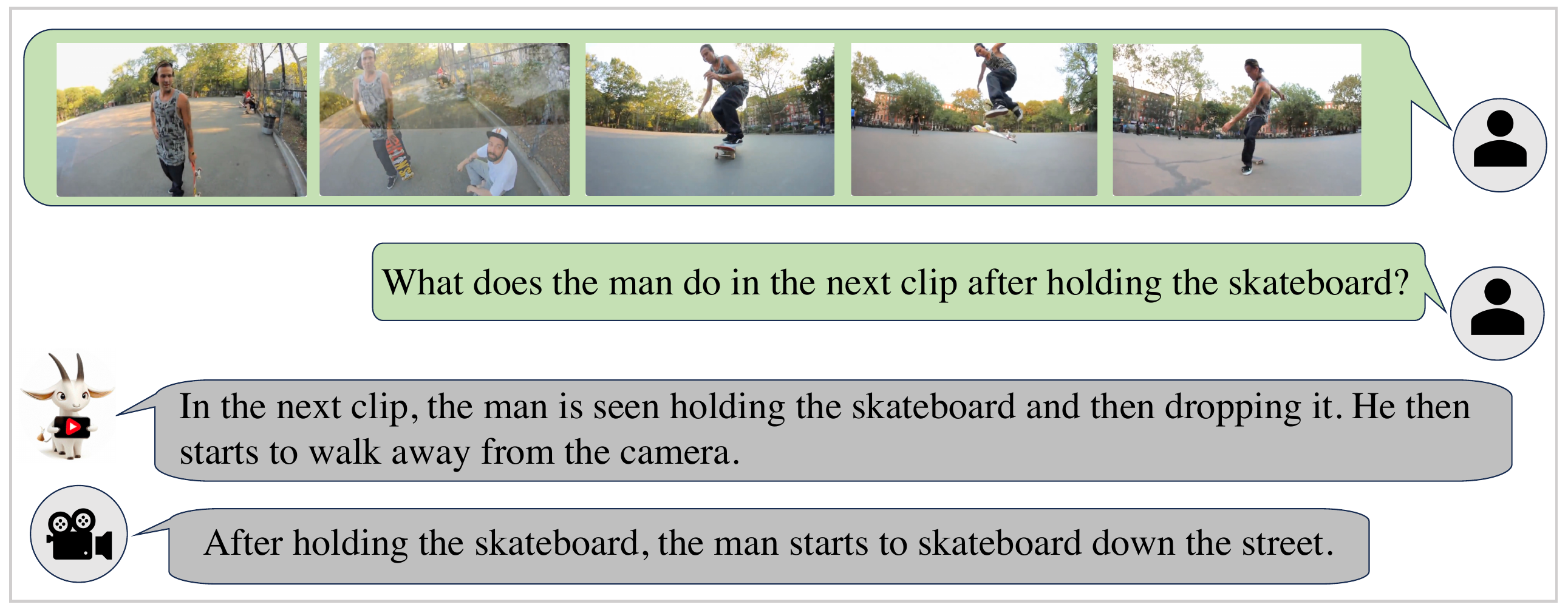} 
    \captionsetup{font={footnotesize}}
    \vspace{-1.8em}
    \caption{Qualitative results of video conversation in terms of actions in a specific frame of the video. We present the answers from VideoChatGPT (upper) and our BT-Adapter-LLaVA (down).}
    \label{visualize22}
    \vspace{-0.5em}
\end{figure}

\begin{figure}[htb] 
    \centering
    \includegraphics[width=1\linewidth]{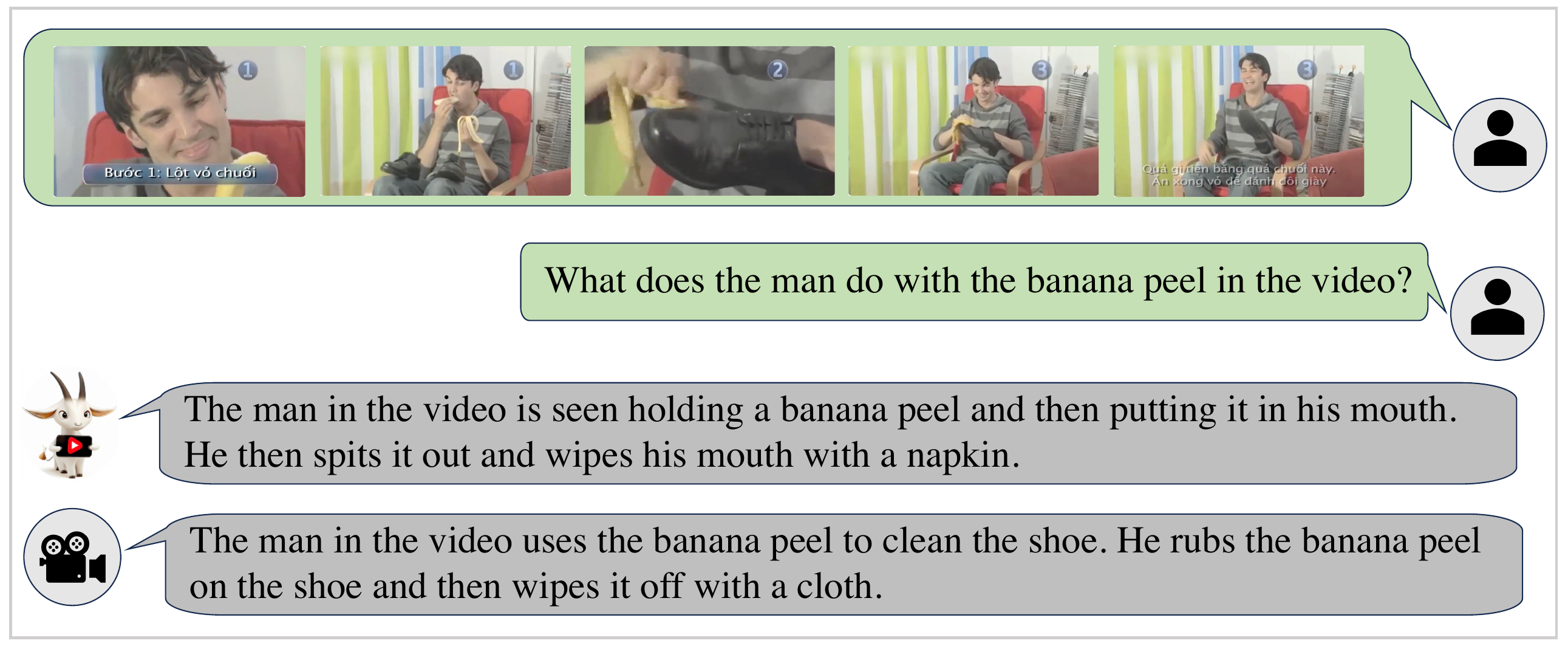} 
    \captionsetup{font={footnotesize}}
    \vspace{-1.8em}
    \caption{Qualitative results of video conversation in terms of unusual actions in the video. We present the answers from VideoChatGPT (upper) and our BT-Adapter-LLaVA (down).}
    \label{visualize33}
    \vspace{-0.5em}
\end{figure}

\begin{figure}[htb] 
    \centering
    \includegraphics[width=1\linewidth]{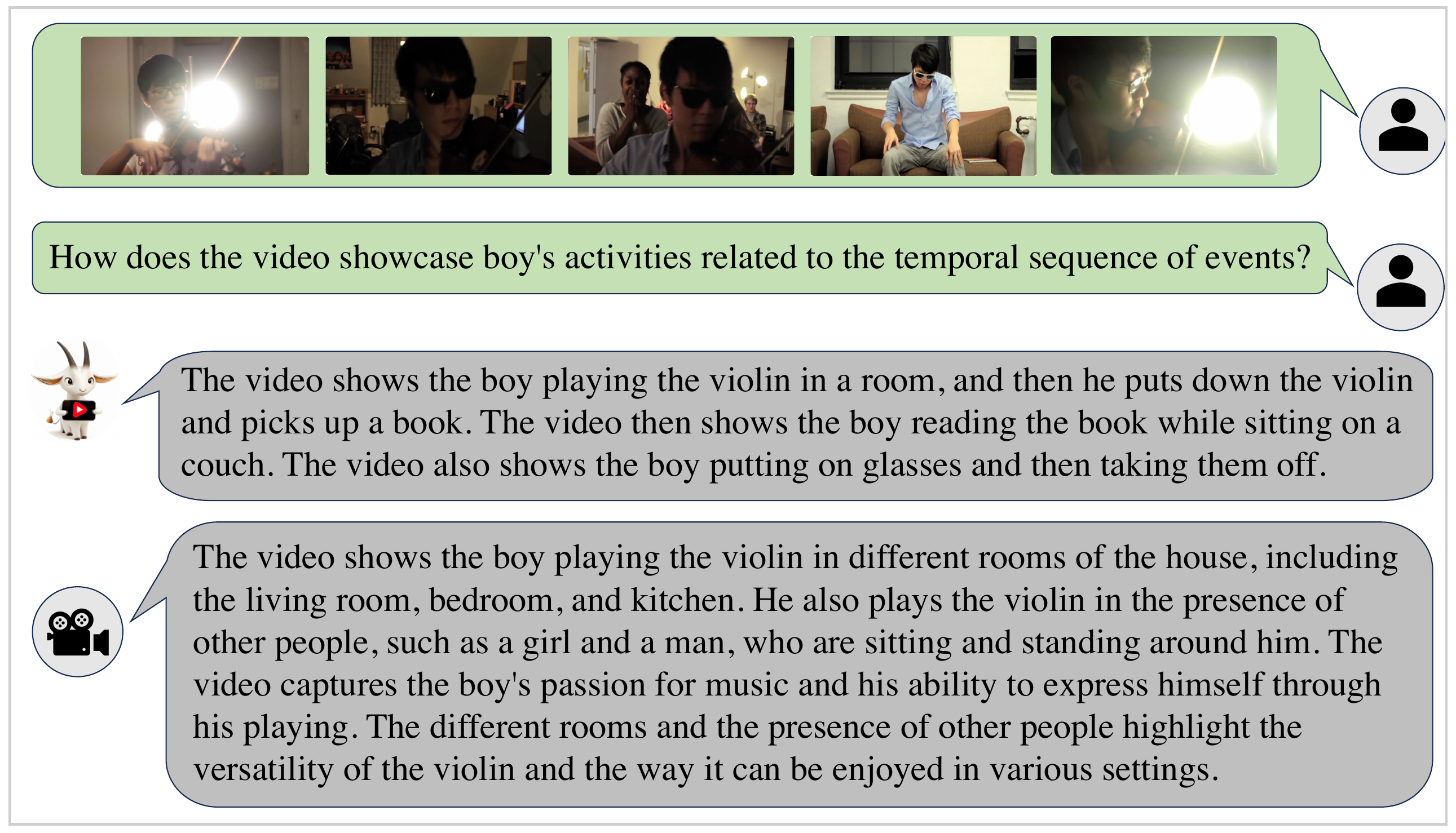} 
    \captionsetup{font={footnotesize}}
    \vspace{-1.8em}
    \caption{Qualitative results of video conversation in terms of complex actions and scenes in a long video (3 min). We present the answers from VideoChatGPT (upper) and our BT-Adapter-LLaVA (down).}
    \label{visualize44}
    \vspace{-0.5em}
\end{figure}

\begin{figure}[htb] 
    \centering
    \includegraphics[width=1\linewidth]{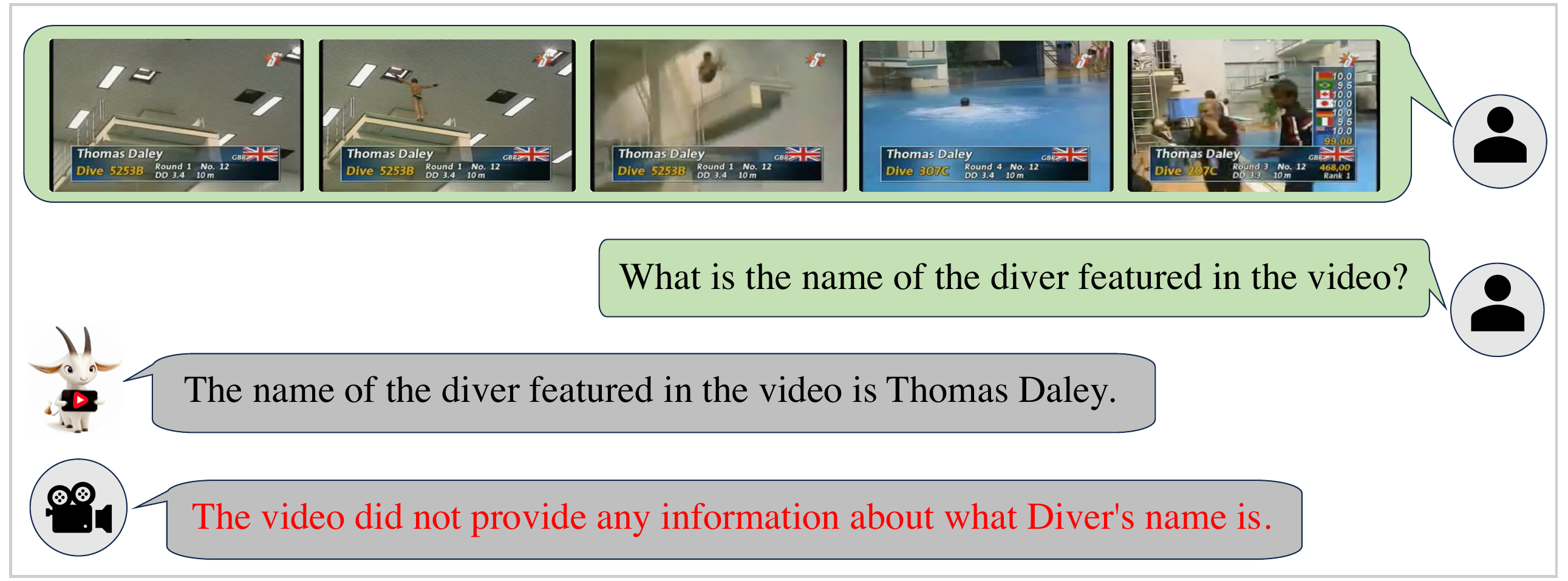} 
    \captionsetup{font={footnotesize}}
    \vspace{-1.8em}
    \caption{The bad case of our method in video conversation. We present the answers from VideoChatGPT (upper) and our BT-Adapter-LLaVA (down).}
    \label{visualize55}
    \vspace{-0.5em}
\end{figure}

\section{More Qualitative Results}
In Figures \ref{visualize11}, \ref{visualize22}, \ref{visualize33}, and \ref{visualize44}, we present a comprehensive overview of the qualitative results obtained in video dialogues, encompassing diverse aspects. These visualizations vividly illustrate the capacity of our BT-Adapter to provide contextually appropriate responses in a variety of scenarios where temporal sensitivity is paramount. These results serve to underscore the efficacy of the BT-Adapter in video understanding. In Figure \ref{visualize55}, we present a notable outlier case in which our method encounters challenges, where the BT-Adapter struggles to recognize the text content within the frames. This particular instance sheds light on the fact that, while the BT-Adapter diligently strives to preserve pretraining information to the greatest extent possible, it may still introduce some disruption to the pretraining knowledge compared to the fully concatenation-based modeling of VideoChatGPT.

\section{Broader Impact}
The research presented in this paper, which leverages Large Language Models (LLMs), comes with several important considerations for its broader impact. The use of LLMs to generate content, while powerful, can inherit biases from the data used for training, potentially resulting in content that reflects these biases, some of which may have negative societal implications. Moreover, the model may generate inaccurate or non-factual content, which can undermine trust in online information sources.

\end{document}